\documentclass[letterpaper]{article} 
\usepackage{aaai23}  
\usepackage{times}  
\usepackage{helvet}  
\usepackage{courier}  
\usepackage[hyphens]{url}  
\usepackage{graphicx} 
\usepackage{natbib}  
\usepackage{caption} 
\usepackage{algorithm}
\usepackage{algorithmic}
\usepackage{enumitem}
\usepackage{multirow}
\usepackage{bbm}
\usepackage{subfigure}
\usepackage{amssymb}
\usepackage{mathtools}
\usepackage{amsthm}
\usepackage{amsmath}
\usepackage{booktabs}
\usepackage{color}
\usepackage{newfloat}
\usepackage{listings}
\DeclareCaptionStyle{ruled}{labelfont=normalfont,labelsep=colon,strut=off} 
\floatstyle{ruled}
\newfloat{listing}{tb}{lst}{}
\floatname{listing}{Listing}
\title{SoftCorrect: Error Correction with Soft Detection for Automatic Speech Recognition}
\author{
      Yichong Leng\textsuperscript{\rm 1}\thanks{This work was conducted at Microsoft. Corresponding author: Xu Tan, xuta@microsoft.com}, Xu Tan\textsuperscript{\rm 2}, Wenjie Liu\textsuperscript{\rm 3}, Kaitao Song\textsuperscript{\rm 2}, Rui Wang\textsuperscript{\rm 2}\\
  \textbf{Xiang-Yang Li\textsuperscript{\rm 1}, Tao Qin\textsuperscript{\rm 2}, Edward Lin\textsuperscript{\rm 3}, Tie-Yan Liu\textsuperscript{\rm 2}}
}
  
\affiliations{
    \textsuperscript{\rm 1}University of Science and Technology of China \textsuperscript{\rm 2}Microsoft Research Asia \textsuperscript{\rm 3}Microsoft Azure Speech\\
    
    \textsuperscript{\rm 1}lyc123go@mail.ustc.edu.cn,xiangyangli@ustc.edu.cn\\
    \textsuperscript{\rm 2}\{xuta,kaitaosong,ruiwa,taoqin,tyliu\}@microsoft.com
    \textsuperscript{\rm 3}\{liwenjie,edlin\}@microsoft.com
}

\begin{document}

\maketitle

\begin{abstract}
    Error correction in automatic speech recognition (ASR) aims to correct those incorrect words in sentences generated by ASR models. Since recent ASR models usually have low word error rate (WER), to avoid affecting originally correct tokens, error correction models should only modify incorrect words, and therefore detecting incorrect words is important for error correction. Previous works on error correction  either \textit{implicitly} detect error words through target-source attention or CTC (connectionist temporal classification) loss, or \textit{explicitly} locate specific deletion/substitution/insertion errors. However, implicit error detection does not provide clear signal about which tokens are incorrect and explicit error detection suffers from low detection accuracy. In this paper, we propose SoftCorrect with a soft error detection mechanism to avoid the limitations of both explicit and implicit error detection. Specifically, we first detect whether a token is correct or not through a probability produced by a dedicatedly designed language model, and then design a constrained CTC loss that only duplicates the detected incorrect tokens to let the decoder focus on the correction of error tokens. Compared with implicit error detection with CTC loss, SoftCorrect provides explicit signal about which words are incorrect and thus does not need to duplicate every token but only incorrect tokens; compared with explicit error detection, SoftCorrect does not detect specific deletion/substitution/insertion errors but just leaves it to CTC loss. Experiments on AISHELL-1 and Aidatatang datasets show that SoftCorrect achieves 26.1\% and 9.4\% CER reduction respectively, outperforming previous works by a large margin, while still enjoying fast speed of parallel generation.
\end{abstract}

\section{Introduction}

\label{sec_intro}

Correction~\cite{cucu2013stat,d2016automatic,anantaram2018repairing,du2022crossmodal} has been widely used in automatic speech recognition (ASR) to refine the output sentences of ASR systems to reduce word error rate (WER). Considering the error rate of the sentences generated by ASR is usually low (e.g., $<$10\%, which means only a small proportion of tokens are incorrect and need correction), how to accurately detect errors is important for correction~\cite{leng2021fastcorrect}. Otherwise, correct tokens may be changed by mistake, or error tokens cannot be corrected. Previous works conduct error detection in different ways: 1) Implicit error detection, where the errors are not explicitly detected but embedded in the correction process. For example, ~\citet{liao2020improving,mani2020asr,wang2020asr,linchen2021improve} adopt an encoder-decoder based autoregressive correction model with a  target-source (decoder-encoder) attention~\cite{vaswani2017attention}; ~\citet{gu-kong-2021-fully} duplicate the source tokens several times and leverage a CTC (connectionist temporal classification) loss~\cite{graves2006connectionist}, where the target-source alignments learnt in decoder-encoder attention or CTC paths play a role of implicit error detection. 2) Explicit error detection, where the specific deletion/substitution/insertion errors are detected out explicitly. For example, \citet{leng2021fastcorrect,leng2021fastcorrect2,du2022crossmodal,shen2022maskcorrect} rely on the predicted duration to determine how many target tokens each source token should be corrected to (e.g., $0$ stands for deletion error, $1$ stands for no change or substitution error, $\ge2$ stands for insertion error).

Implicit error detection enjoys the advantage of the flexibility of model learning but suffers from the limitation that it does not provide clear signal for model training about which tokens are incorrect. In the contrast, explicit error detection enjoys the advantage of clear signal but suffers from the limitation that it requires precise error patterns and thus new error will be introduced once error detection is not accurate. For example, if a substitution error is predicted as an insertion error by explicit error detection, then the model cannot correct this error but introduce new error by inserting a wrong token. A natural question arises: can we design a better error detection mechanism that inherits the advantages of both implicit and explicit error detection and avoids their limitations? 

To answer this question, in this paper, we propose a soft error detection mechanism with an error detector (encoder) and an error corrector (decoder). Specifically, we dedicatedly design a language model as the encoder to determine whether a token is correct or not and design a constrained CTC loss on the decoder to only focus on correcting the detected error tokens (focused error detection):

\begin{itemize}[leftmargin=*]
\item Instead of predicting correction operations such as
deletion, substitution and insertion~\cite{leng2021fastcorrect,leng2021fastcorrect2,du2022crossmodal}, we only detect whether a token is correct or not. To this end, we can either use a binary classification \cite{fang-etal-2022-non} or the probability from language model for error detection. We choose the latter one since  the probability from a language model contains more knowledge in language understanding and vocabulary space \cite{hinton2015distilling,gou2021knowledge} than a simple binary classification. However, previous methods for language modeling such as left-to-right language modeling (e.g., GPT~\cite{Radford2018GPT,radford2019GPT2,Brown2020GPT3}) and bidirectional language modeling (e.g., BERT~\cite{devlin2019bert}) are not suitable in our scenario: 1) left-to-right language models like GPT only leverage unidirectional context information (e.g., left), which cannot provide enough context information for accurate probability estimation; 2)  bidirectional language models like BERT can leverage bidirectional context information, but it needs N passes (where N corresponds to the number of tokens)~\cite{Julian2020Masked} to estimate the probability of all the $N$ tokens in a sentence, which cannot satisfy the fast speed requirement for error detection. In this paper, we train the encoder with a novel language model loss (see \textsection\ref{sec_method_enc_det}), to output probabilities effectively and efficiently to detect error tokens in source sentence.

\item Instead of duplicating all the source tokens multiple times in CTC loss, we only duplicate the incorrect tokens detected (as indicated by the probabilities from the encoder trained with our novel language model loss) and use a constrained CTC loss to let the decoder focus on the correction of these duplicated error tokens, resulting in a focused error correction. Compared with the standard CTC loss that duplicates all the tokens, our constrained CTC loss provides clear signals about which part of tokens should be corrected. 
\end{itemize}

Furthermore, previous works \cite{weng2020joint,Yuchen2018Reranking} have shown that the multiple candidates generated by ASR beam search can be leveraged to verify the correctness of tokens \cite{leng2021fastcorrect2} in each candidate. To further improve correction accuracy, we take multiple candidates from ASR beam search as encoder input. Accordingly, the error detection in the encoder contains two steps, i.e., first selecting a better candidate from multiple candidates (equivalent to detect which candidates are likely to be incorrect) for further correction, and then detecting which tokens are likely to be incorrect in the selected candidate. The contributions of this paper are summarized as follows:
\begin{itemize}[leftmargin=*]

\item We propose SoftCorrect with a soft error detection mechanism for ASR error correction to inherit the advantages of both explicit and implicit error detection and avoid their limitations.
\item We design a novel language model loss for encoder to enable error detection and a constrained CTC loss for the decoder to focus on the tokens that are detected as errors.
\item Experimental results on AISHELL-1 and Aidatatang datasets demonstrate that SoftCorrect achieves 26.1\% and 9.4\% CER reduction respectively, while still enjoying fast error correction with parallel generation.
\end{itemize}

The source code of SoftCorrect will be available in https://github.com/microsoft/NeuralSpeech.

\section{Background}

\subsection{Error Correction for ASR} 
\paragraph{Error Correction Models} 
Error correction is widely used in ASR systems~\cite{shivakumar2019learning,hu2020deliberation} to reduce word error rate. Error correction models usually take the sentences outputted by ASR systems as input and generate corrected sentences, and have evolved from early statistic machine translation models~\cite{cucu2013stat,d2016automatic}, to later neural-network based autoregressive models~\cite{tanaka2018neural,liao2020improving,wang2020asr}, and to recent non-autoregressive models~\cite{leng2021fastcorrect,leng2021fastcorrect2,du2022crossmodal}. Non-autoregressive error correction models generate sentences in parallel with the help of a duration predictor~\cite{Jiatao2018NAT} to predict the number of tokens that each input token can be corrected to, which achieve much faster inference speed than autoregressive counterparts and approximate correction accuracy, making it suitable for online deployment.

\label{Voting}
\paragraph{Multiple Candidates} Recent works~\cite{linchen2021improve,Yuchen2018Reranking,Imamura2017Reranking,weng2020joint} show that the multiple candidates generated by ASR beam search can have \textit{voting effect}~\cite{leng2021fastcorrect2}, which can be beneficial for both autoregressive and non-autoregressive correction models. They first align the multiple candidates to the same length using multi-candidate alignment algorithm based on the token-level similarity and phoneme-level similarity,
and take the aligned candidates as encoder input. SoftCorrect also leverages multiple candidates since the difference of beam search results can show the uncertainty of ASR model and give clues about potential error tokens.

\subsection{Error Detection by Target-Source Alignments}
Error detection can be achieved via the alignments between the target (correct) sentence and the source (incorrect) sentence, which can be either explicit or implicit. 

\label{para_explicit}
\paragraph{Explicit Alignment} By explicitly aligning the source and target sequences together with edit distance~\cite{leng2021fastcorrect}, we can obtain the number of target tokens (duration) aligned with each source token and train a duration predictor. Thus, we can detect insertion, deletion and substitution error with corresponding duration (e.g., $0$ stands for deletion error, $1$ stands for no change or substitution error, $\ge2$ stands for insertion error). However, the duration predictor is hard to optimize precisely and thus new error will be introduced once duration prediction is not accurate. 

\begin{figure*}[!t]
    \centering
    \includegraphics[width=0.80\textwidth]{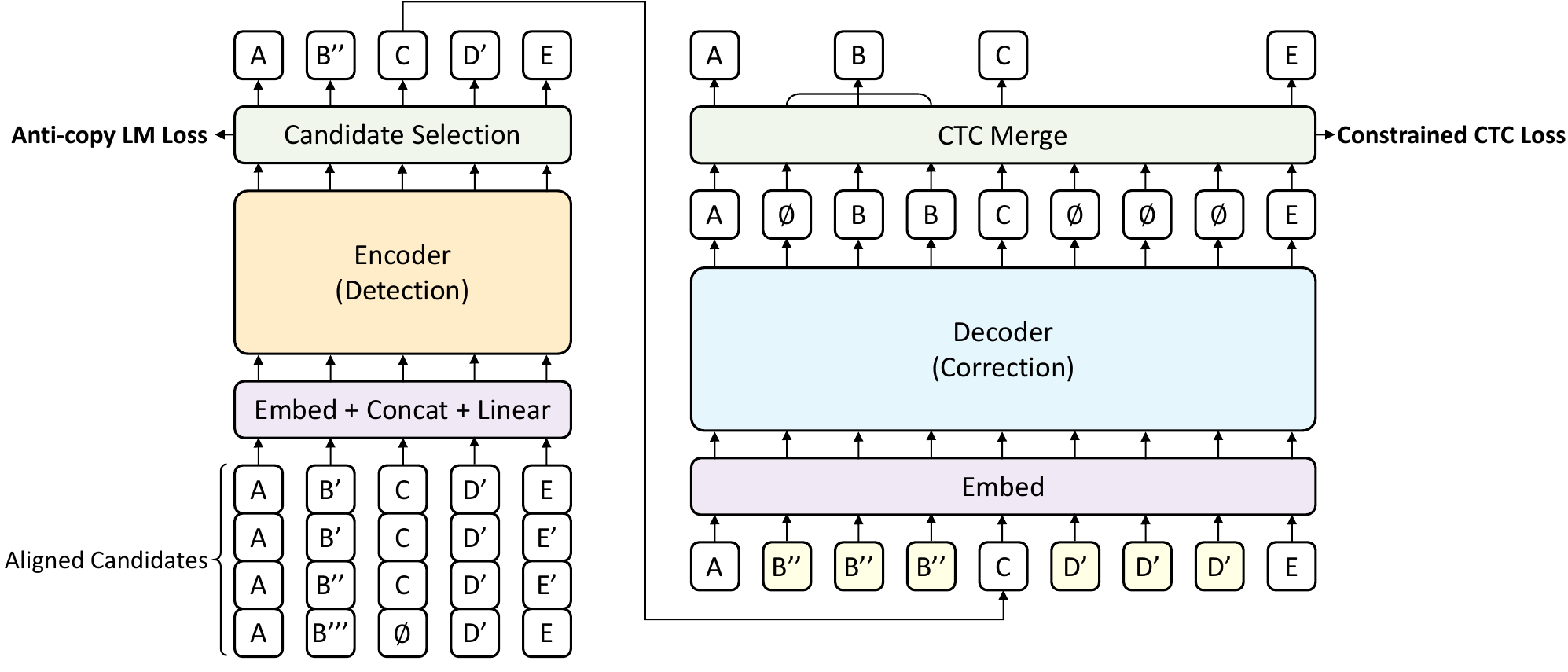}

    \caption{Overview of SoftCorrect. We use $A~B~C~E$ to represent the ground-truth tokens, while $B'~B''~B'''~D'~E'$ to represent incorrect tokens. We use $\phi$ to represent blank token for alignment purpose only, which is leveraged in both multi-candidate alignment and CTC alignment. In this case, the ground-truth sentence is $ABCE$, while the $4$ candidates are $AB'CD'E$, $AB'CD'E'$, $AB''CD'E'$, and $AB'''D'E$, respectively. The selected candidate is $AB''CD'E$, where $B''$ and $D'$ are detected as incorrect tokens and duplicated when fed into decoder.}
    \label{fig_system_overview}
    \vspace{-3pt}
\end{figure*}

\paragraph{Implicit Alignment} Errors can be ``detected'' via implicit alignment between target and source sequences. For example, Transformer \cite{vaswani2017attention} based autoregressive models embed the target-source alignment in decoder-encoder attention \cite{wang2020asr,linchen2021improve}, and CTC-based models \cite{libovicky2018end,saharia2020non,majumdar2021citrinet} leverage a CTC loss~\cite{graves2006connectionist} to align target with duplicated source implicitly. A desirable property of CTC loss is that it enables parallel error correction, without the need of duration prediction and with more flexibility during correction. Thus, in this paper, we adopt the CTC based solution but enhance it with a soft error detection mechanism.

\section{SoftCorrect}

\subsection{System Overview}
\label{sec:overview}
As shown in Figure~\ref{fig_system_overview}, SoftCorrect consists of an error detector (the encoder) and a focused error corrector (the decoder). We introduce the whole system step by step:
\begin{itemize}[leftmargin=*]
\item Motivated by the voting effect
in multiple candidates for error detection, we leverage multiple candidates from ASR beam search, as introduced in \textsection\ref{Voting}. We first align these candidates to the same length following~\citet{leng2021fastcorrect2}. The aligned candidates are shown in the bottom left of Figure~\ref{fig_system_overview}. The aligned candidates are converted into token embeddings, concatenated along the position and fed into a linear layer.

\item The detector is a standard Transformer Encoder~\cite{vaswani2017attention} which takes the output of the previous step as input, and generates a probability for each token in each candidate. Specifically, the output hidden of the encoder is multiplied with a token-embedding matrix to generate a probability distribution over the whole vocabulary.
For example, the output probability distribution in the last position in Figure~\ref{fig_system_overview} can provide the probability for token $E$ and $E'$ simultaneously. Since ASR usually has low WER, to prevent encoder from learning trivial copy, we propose an anti-copy language model loss to train the encoder to output this probability distribution, as introduced in \textsection\ref{sec_method_enc_det}. 

\item Based on the probability, we can choose the token with the highest probability in each position from multiple candidates and obtain a better candidate, which usually contains less errors and thus makes the error detection easier.
This step is illustrated as the ``Candidate Selection'' module in Figure~\ref{fig_system_overview} and the selected candidate is $AB''CD'E$.
Noted that we conduct \textit{position-wise} selection and the tokens in selected candidate (e.g., $B''$ and $E$) can come from different candidates.
\item After candidate selection, we combine the probability of each token in selected candidate with its corresponding probability from the ASR model. 
{The error detection score for each token is the weighted linear combination of encoder probability and ASR output probability reflecting the similarity between token pronunciation and audio.}
A token is detected as incorrect when the combined probability is lower than a threshold \cite{huang2019empirical}. 
\item The corrector (decoder) takes the generated candidate as input and outputs refined tokens. It is trained with a constrained CTC loss as introduced in \textsection\ref{sec_method_dec_cor}, which learns to only modify the detected ``incorrect” tokens while directly copying remaining tokens to output. Therefore, we only duplicate the incorrect tokens detected in previous step. As shown in the bottom right of Figure~\ref{fig_system_overview}, the detected and duplicated error tokens are $B''$ and $D'$. 
\end{itemize}

In the next subsections, we introduce the details of the anti-copy language model loss in encoder to generate token probability for error detection, and the constrained CTC loss in decoder for focused error correction. {More discussion of SoftCorrect about the related works, limitations and future works can be found in Appendix \ref{appen:further_dis}.}

\subsection{Anti-Copy Language Modeling for Detection}
\label{sec_method_enc_det}

\begin{figure*}[!t]
    \centering
    \includegraphics[width=0.66\textwidth]{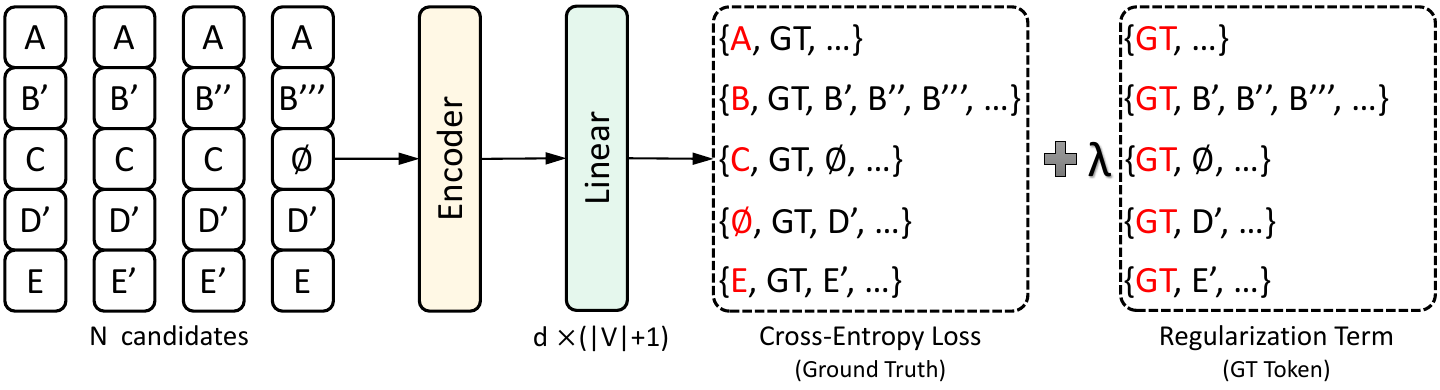}
    \caption{Illustration of the proposed anti-copy language model loss as formulated in Equation~\ref{eq_lm_loss}. The tokens in red color represent the target tokens in each term. The remaining tokens in the vocabulary are represented as ``...''.}
    \label{fig_lm_loss}
    \vspace{-3mm}
\end{figure*}

We use the encoder to output a probability for each token in the multiple candidates, where this probability can be used in candidate selection and error detection.
Since we need to select better candidate as well as detect errors, the probability from binary classification based error detection method~\cite{omelianchuk-etal-2020-gector} is unsuitable for the lack of knowledge in language understanding and vocabulary space \cite{hinton2015distilling,gou2021knowledge}, which is also verified empirically in \textsection\ref{sec_ablation}. Indeed, the probability we need is like a kind of language modeling, which determines whether a token is natural/reasonable given the context information. 
{As discussed in Section \ref{sec_intro}, common language modeling methods are unsuitable for probability estimation since GPT-based methods lack bidirectional context information and BERT-based methods are too slow with N-pass~\cite{Julian2020Masked} inference.
To alleviate these issues, we propose a novel method to not only leverage bidirectional information but also provide fast probability estimation.

A straightforward way is to train the Transformer encoder to predict ground-truth (correct) tokens in each position given multiple aligned candidates as input. In this way, the encoder can learn to output probability to determine whether a token is natural/reasonable given the context information. However, since the ASR systems usually have relatively low WER (e.g., $<$10\%), a \textit{large proportion} of tokens in aligned input are consistent (i.e., with the same token) and correct. In this way, directly predicting the ground-truth token in each position would result in trivial copy. For example, the first position in Figure \ref{fig_system_overview} might be a trivial copy since the aligned input tokens are consistent (i.e., all $A$) and the corresponding target token is also $A$. This trivial copy issue will cause the model outputting an extreme high probability for the input token on consistent position and thus hurt the ability of encoder on detecting errors on consistent position.

To alleviate this problem, we propose an anti-copy language model loss to prevent learning copy only, which modifies the standard cross-entropy loss by changing its prediction vocabulary and adding a regularization term. Specifically, we add a special $GT$ token (it does not stand for any specific ground-truth token in each position, but just a special symbol) in the vocabulary as shown in Figure~\ref{fig_lm_loss}, and the objective function to train the encoder is as follows. 
\begin{equation}
\begin{aligned}
 \mathcal{L}_{\rm lm} &= \sum^{N}_{t=1} \frac{\exp (H_{t}W_{y_t})} {\sum_{i \in \{V+ GT\}} \exp (H_{t}W_i)} \\
 &+ \lambda \sum^{N}_{t=1} \frac{\exp (H_{t}W_{GT})} {\sum_{i \in \{V\setminus y_t + GT\}} \exp (H_{t}W_i)},
\end{aligned}
\label{eq_lm_loss}
\end{equation}
where $V$ represents the original token vocabulary (including a special token $\phi$ to represent deletion), $V+GT$ represents the original token vocabulary plus $GT$ token, and $V\setminus y_t$ represents the original token vocabulary minus $y_t$, where $y_t$ is the ground-truth token at position $t$. $W \in \mathbb{R}^{d\times (|V|+1)}$, $H \in \mathbb{R}^{N\times d}$, where $N$ is the length of aligned candidates and $d$ is the hidden size of the encoder output. $W_{y_t} \in \mathbb{R}^{d}$ and $W_{GT} \in \mathbb{R}^{d}$ represent the vector in the softmax matrix that corresponds to token $y_t$ and $GT$ respectively, and $H_t$ represents the hidden vector generated by the encoder at position $t$. $\lambda$ is used to balance the regularization term.

The first term in Equation~\ref{eq_lm_loss} is a cross-entropy loss to predict the ground-truth token over the full vocabulary (including the $GT$ token). The second is a regularization term, which predicts $GT$ token over the full vocabulary without ground-truth token (including the $GT$ token but removing the ground-truth token). The first term is used to encourage the encoder to output ground-truth token for reasonable probability estimation, and the second term is used to alleviate copying the ground-truth token.

There are two intuitions behind the anti-copy loss: 1) The regularization (second) term of anti-copy loss aims to avoid copy-mapping by removing ground-truth token from vocabulary and train the model to predict $GT$ over all vocabulary except the ground-truth token; 2) Since all error tokens are optimized to have lower probability than $GT$ while the correct token is optimized to have higher probability than $GT$, an advantages of anti-copy loss is that the $GT$ token can serve as a decision boundary between correct token and error token and help better detect the error token based on the probability of language model with anti-copy loss.}

\subsection{Constrained CTC Loss for Correction}
\label{sec_method_dec_cor}

\begin{figure}[!t]
    \centering
    \includegraphics[width=0.45\textwidth]{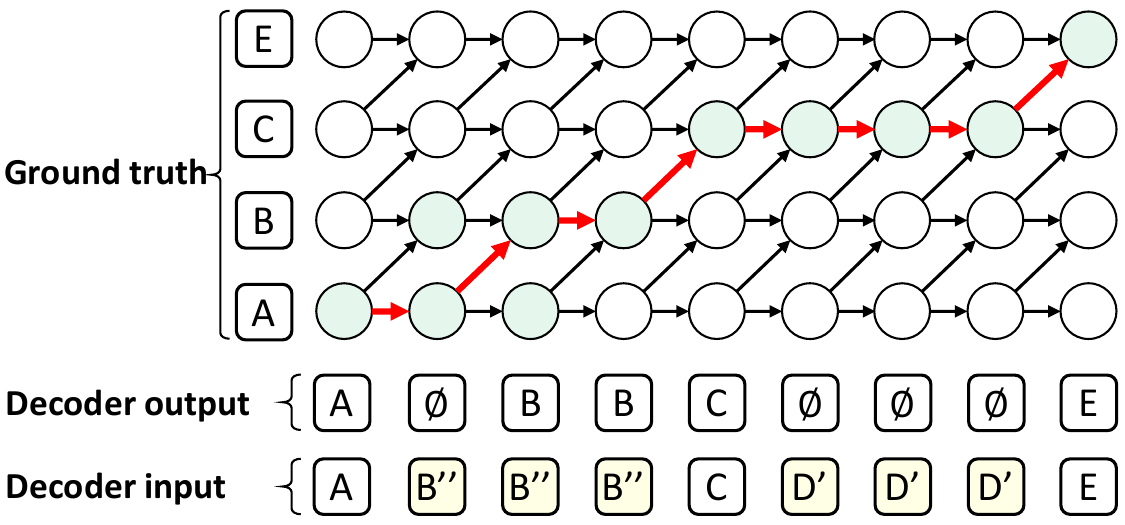}
    \caption{Constrained CTC loss, which only allows dynamic alignment between the output and target for the detected (repeated) tokens (e.g., $B''$ and $D'$), but uses fixed alignment for the undetected (not repeated) tokens (e.g., $A$, $C$, and $E$). The nodes in green color represent all possible CTC alignments while the nodes connected with red arrows represent the alignment for the decoder output (just one possible case) shown in this figure.}
    \vspace{-3mm}
    \label{fig_ctc}
\end{figure}

As aforementioned, we obtain the constructed candidate from the encoder with detected errors and feed it as the input of the decoder to generate the final corrected result. 
Since we know which tokens in the selected candidate are correct or incorrect, it is unnecessary to modify all the tokens in the decoder, which can cause larger latency and produce new errors (e.g., a correct token is edited as an incorrect token). Therefore, we propose a constrained CTC loss as shown in Figure~\ref{fig_ctc}. We only repeat the incorrect tokens and keep the remaining tokens unchanged for decoder input. Incorrect tokens are repeated three times, which is a common practice for CTC loss~\cite{libovicky2018end,saharia2020non}. The likelihood of the target sequence $y$ is the probability summation of all possible CTC alignment paths after constraints:
\begin{equation}
 \mathcal{L}_{\rm ctc} = \sum_{z \in \phi'(y)}{P(z|x)},
    \label{eq_ctc_loss}
\end{equation}
where $x$ is the decoder input (i.e., the selected candidate with expansion on the ``incorrect tokens'', just as ``decoder input'' shown in Figure~\ref{fig_ctc}), 
$\phi'(y)$ represents all the possible alignment paths to generate $y$ in our constrained CTC (different from the standard alignments $\phi(y)$), and $z$ represents one possible CTC path, as shown in the red arrow in Figure~\ref{fig_ctc}.

The main difference between our loss and standard CTC loss is the \textit{all possible alignment paths} $\phi'(y)$. As shown in Figure~\ref{fig_ctc}, the possible CTC alignment paths $\phi'(y)$ lay only on the shaded (green) nodes, in contrast to the standard CTC where the alignment paths $\phi(y)$ lay on all nodes. As a result, the ``correct token'' is used as an anchor and cannot be dynamically aligned (the output of anchor token must be the anchor token itself). {During the inference, we perform softmax over the possible error tokens, select the best token from each of those positions and then remove duplicates and blank.} With the help of the explicit error detection from encoder, we can skip the correction process of the decoder to reduce the system latency if all input tokens are detected as correct token.

{Furthermore, as discussed in Section~\ref{para_explicit}, explicit error detector can possibly produce new errors (e.g., some correct tokens are identified as errors) and propagate them to the error corrector. Hence, we need the error corrector to be more robust to the outputs of error detector. Specifically, when training corrector, we randomly select 5\% correct tokens and regard them as pseudo error tokens to simulate the mistakes from detector, so that model will not modify these correct tokens during the optimization and can be more robust to the outputs of detector. More discussion about this design is in Appendix \ref{appen:further_dis}.}

\section{Experimental Setup}
\label{sec:exp_setup}
In this section, we introduce the datasets and the ASR model used for correction, and some previous error correction systems for comparison.

\subsection{Datasets and ASR Model}
We conduct experiments on two Mandarin ASR datasets, AISHELL-1~\cite{bu2017aishell} and Aidatatang\_200zh ~\cite{aidatatang}, referred to as Aidatatang for short. AISHELL-1 contains 150/10/5-hour speech data for train/ development/test, while Aidatatang contains 140/20/40-hour speech data for train/development/test, respectively.

The ASR model used in our experiments is a state-of-the-art model with Conformer architecture~\cite{gulati2020conformer}, enhanced with SpecAugment~\cite{park2019specaugment} and speed perturbation for data augmentation, and a language model for joint decoding. The hyper-parameters of this ASR model follow the ESPnet codebase~\cite{watanabe2018espnet}\footnote{\small github.com/espnet/espnet/tree/master/egs/aishell}\footnote{github.com/espnet/espnet/tree/master/egs/aidatatang\_200zh}.

The training, development, and test data for correction models are obtained by using the ASR model to transcribe the corresponding datasets in AISHELL-1 and Aidatatang. Following the common practice in ASR correction \cite{leng2021fastcorrect,leng2021fastcorrect2,du2022crossmodal,linchen2021improve}, we use 400M unpaired text data to construct a pseudo pretraining dataset for both SoftCorrect and baseline systems. The details of pseudo data generation can be found in Appendix \ref{appen:data_gen}.

\subsection{Baseline Systems}
We compare SoftCorrect with the several error correction baselines, including systems using implicit and explicit error detection.

For baselines with implicit error detection, we use: 1) \textit{AR Correct}. A standard autoregressive (AR) encoder-decoder model based on Transformer~\cite{vaswani2017attention}. 2) \textit{AR N-Best}. Following \citet{linchen2021improve}, we train an AR Transformer model by taking the aligned multiple candidates from ASR beam search as input and generating correction result.

For baselines with explicit error detection, we use: 1) \textit{FastCorrect}. FastCorrect~\citep{leng2021fastcorrect} is a non-autoregressive model for ASR correction, which utilizes token duration to adjust input sentence length to enable parallel decoding. 2) \textit{FastCorrect 2}. \citet{leng2021fastcorrect2} introduce multiple candidates into non-autoregressive FastCorrect model and achieve state-of-the-art correction accuracy.

Considering SoftCorrect leverages multiple candidates, we also take rescoring method for comparison. The rescoring model is a 12-layer Transformer decoder model and the details of rescoring follow \citet{huang2019empirical}. Besides, we also combine non-autoregressive correction with rescoring together to construct another two baselines. One is \textit{FC + Rescore} where the outputs of ASR are first corrected by FastCorrect (FC for short) and then rescored, the other is \textit{Rescore + FC} where the ASR outputs are first rescored and then corrected.

\begin{table*}[t]
\caption{The correction accuracy and inference latency of different systems. We report the character error rate (CER) and character error rate reduction (CERR) on test and development sets of the two datasets, and report the inference latency measured on NVIDIA V100 GPU or "Intel(R) Xeon(R) Platinum 8168 CPU @ 2.70GHz" CPU on the test set of AISHELL-1. ``FC'' stands for FastCorrect in other baselines.}
\vspace{-2mm}
\label{tab:main_result}
\begin{small}
\begin{center}
\scalebox{0.84}{
\begin{tabular}{l|c|c|c|c|c|c|c|c|c|c}
\toprule
\multirow{3}{*}{Model} & \multicolumn{4}{c|}{AISHELL-1}  & \multicolumn{4}{c|}{Aidatatang} & \multicolumn{2}{c}{\multirow{2}{*}{\shortstack{ Latency \\ (ms/sent)}}} \\
\cmidrule{2-9} 
& \multicolumn{2}{c|}{Test}  & \multicolumn{2}{c|}{Dev}  &  \multicolumn{2}{c|}{Test}   & \multicolumn{2}{c|}{Dev} & \multicolumn{2}{}{} \\
\cmidrule{2-11} 
& CER  & CERR & CER & CERR & CER & CERR & CER  & CERR & GPU& CPU \\
\midrule
No Correction & 4.83 & - & 4.46 & - & 4.47 & -&  3.82    &  -  & - & - \\
\midrule
\multicolumn{10}{l}{\textit{Implicit error detection baselines}}\\\midrule
AR Correct  & 4.07 & 15.73 & 3.79 & 15.02  & 4.39 & 1.79   & 3.74 & 2.09  & 119.0 \small{(1.0$\times$)} & 485.5 \small{(1.0$\times$)} \\
AR N-Best & 3.94 & 18.43 & 3.68 & 17.49  & 4.70 & -5.15 & 4.06 & -6.28 & 121.6 \small{(1.0$\times$)} & 495.8 \small{(1.0$\times$)} \\
\midrule
\multicolumn{10}{l}{\textit{Explicit error detection baselines}}\\\midrule
 FastCorrect& 4.16 & 13.87 & 3.89 & 12.78 & 4.47 & 0.00  & 3.82 & 0.00  & 16.2 \small{(7.2$\times$)} & 92.1 \small{(5.3$\times$)} \\
FastCorrect 2& 4.11 & 14.91& 3.78 & 15.25 & 4.59 & -2.68 & 4.02 & -5.24  & 23.1 \small{(5.2$\times$)}  & 114.6 \small{(4.2$\times$)}   \\
\midrule
\multicolumn{10}{l}{\textit{Other baselines}}\\\midrule
Rescore & 4.02 & 16.77 & 3.74 & 16.14 & 4.29 & 4.03 & 3.64 & 4.71 & 48.8 \small{(2.4$\times$)}  & 256.0 \small{(1.9$\times$)}  \\
Rescore + FC & 3.69 & 23.60 & 3.48 & 21.97 & 4.33 & 3.13 & 3.68 & 3.66 & 65.0 \small{(1.8$\times$)}  & 348.1 \small{(1.4$\times$)}  \\
FC + Rescore & 3.58& 25.88 & 3.40 & 23.77 & 4.29 & 4.03 & 3.63 & 4.97 & 113.6 \small{(1.0$\times$)}  & 624.4 \small{(0.8$\times$)}  \\\midrule
SoftCorrect  & \textbf{3.57} & \textbf{26.09}  & \textbf{3.40} & \textbf{23.77} & \textbf{4.05} & \textbf{9.40} & \textbf{3.44} & \textbf{9.95} & 17.0 \small{(7.0$\times$)}  & 97.4 \small{(5.0$\times$)}  \\
\bottomrule
\end{tabular}
}
\end{center}
\end{small}
\end{table*}

The detailed training configurations can be found in Appendix \ref{appen:training_details}.

\section{Results}
In this section, we first compare the character error rate reduction (CERR) and latency of SoftCorrect with baselines, and then conduct ablation studies to verify the effectiveness of several designs in SoftCorrect, including anti-copy language model loss and constrained CTC loss. Besides, we conduct some analyses to show the advantages of SoftCorrect on the ability of error detection and error correction over baseline systems.

\subsection{Accuracy and Latency}
\label{subsec:acc_latency}
We report the correction accuracy and inference latency of different systems in Table \ref{tab:main_result}. We have several observations: 

1) Compared with non-autoregressive baselines FastCorrect and FastCorrect 2 with explicit error detection, SoftCorrect achieves larger CERR while still enjoying low latency, which demonstrates the effectiveness of SoftCorrect with soft error detection. 

2) Compared with autoregressive baselines AR and AR N-Best with implicit error detection, SoftCorrect is the first non-autoregressive system achieving larger CERR than them. Later discussion in Section \ref{sec_ablation} about Setting 6 in Table \ref{tab:ablation} shows that SoftCorrect is also better than non-autoregressive baselines with implicit error detection.

3) Compared with combined systems (Rescore+FC and FC+Rescore), SoftCorrect achieves slightly better CERR but much faster inference speed (7x speedup over FC + Rescore in terms of latency). 

4) On Aidatatang dataset, the errors are hard to detect and some previous correction baselines fail to reduce CER (see \textsection\ref{PR_ana} for the analyses on error detection ability), while SoftCorrect still achieves more than 9\% CERR, demonstrating the advantage of our soft detection mechanism by first detecting error tokens and then focusing on correcting errors tokens. Since ASR results have few error tokens, accurate error detection is necessary to avoid mistaking originally correct tokens or missing originally incorrect tokens. As a result, our design of soft error detection achieves good results for ASR correction.

Results on more datasets and significance tests can be found in  Appendix \ref{appen:wenetspeech} and \ref{appen:sig_test}. We also explore the accuracy of SoftCorrect when the CER of ASR system is high in Appendix \ref{appen:p_wrt_wet}.

\begin{table*}[t]
\caption{Ablation studies on the designs in SoftCorrect, including some variants of the anti-copy language model loss for encoder and the constrained CTC loss for decoder. ``CE loss'' means using standard cross-entropy loss to predict ground truth, ``BERT-style'' refers to using BERT model to estimate the probability of each token via N-pass~\cite{Julian2020Masked}. ``GPT-style'' refers to using left-to-right language model to estimate the probability of each token and ``Binary'' refers to using binary classification to detect errors.}
\vspace{-3mm}
\label{tab:ablation}
\begin{small}
\begin{center}
\scalebox{0.88}{
\begin{tabular}{l|c|c|c|c|c|c|c|c|c|c}
\toprule
\multirow{3}{*}{ID} & \multicolumn{2}{|c|}{\multirow{3}{*}{Setting}} & \multicolumn{4}{c|}{AISHELL-1}  & \multicolumn{4}{c}{Aidatatang} \\
\cmidrule{4-11} 
& \multicolumn{2}{c|}{} & \multicolumn{2}{c|}{Test}  & \multicolumn{2}{c|}{Dev}  &  \multicolumn{2}{c|}{Test}   & \multicolumn{2}{c}{Dev} \\
\cmidrule{4-11} 
& \multicolumn{2}{c|}{} & CER $\downarrow$  & CERR $\uparrow$  & CER $\downarrow$  & CERR $\uparrow$ & CER$\downarrow$  & CERR $\uparrow$ & CER$\downarrow$   & CERR $\uparrow$\\
\midrule
1 & \multicolumn{2}{c|}{SoftCorrect} & 3.57 & 26.09 & 3.40 &23.77 & 4.05& 9.40 & 3.44 & 9.95 \\
\midrule
2 & \multirow{5}{*}{Encoder} & CE loss & 3.77 & 21.94 & 3.60 & 19.28 & 4.21 & 5.82 & 3.56 & 6.81 \\
3 & & BERT-style & 3.93  & 18.63 & 3.73 & 16.37  & 4.14 & 7.38 & 3.52 & 7.85 \\
4 & & GPT-style & 4.76  & 1.45& 4.36 & 2.24  & 4.41 & 1.34 & 3.78 & 1.05 \\
5 & & Binary & 3.98 & 17.60 & 3.75 & 15.92  & 4.26 & 4.70 & 3.62 & 5.24 \\
\midrule
6 & Decoder & - Constraint & 3.95 & 18.22 & 3.69 & 17.26 & 4.14 & 7.38 & 3.52 & 7.85 \\
\midrule
7 & \multicolumn{2}{c|}{No Correction} &  4.83  & - &4.46 & -  &  4.47  & - &3.82 & - \\
\bottomrule
\end{tabular}}
\end{center}
\end{small}
\end{table*}

\subsection{Ablation Studies}
\label{sec_ablation}
We conduct ablation studies to verify the effectiveness of our soft detection with encoder for error detection and decoder for focused correction, as shown in Table~\ref{tab:ablation}. We introduce these studies as follows: 

\begin{itemize}[leftmargin=*]
\item We first remove our anti-copy language model loss and simply use standard cross-entropy loss to train the encoder (Setting ID 2), resulting in an inferior accuracy due to the model may learn to copy the ground-truth token, hurting the error detection ability.

\item Since there are some popular methods to model token-level probability such as BERT \cite{devlin2019bert} and GPT \cite{Radford2018GPT}, we apply BERT-style and GPT-style training loss on the encoder to detect errors in the aligned multiple candidates, as shown in ID 3 and 4. Moreover, we also train the encoder to perform detection on each token of each candidate with a binary classification loss (ID 5). The poor results of GPT-style training shows that bi-directional information is necessary for error-detection. The BERT-style training or binary classification achieves lower accuracy than SoftCorrect, showing the effectiveness of our anti-copy language model loss. More details about leveraging the BERT-style and GPT-style loss for detection can be found in Appendix \ref{appen:enc_ablation}.

\item When removing the constraint on the CTC loss (ID 6), the correction accuracy is lower, which demonstrates the advantage of only focusing on correcting the detected error tokens. The results also show that the error detector is reliable because the attempt on modifying tokens that are detected to be right (Setting ID 6) only leads to worse accuracy.
\end{itemize}

\begin{table*}[t]
\small
\caption{Comparison of different systems in terms of error detection and correction ability. $P_{det}$, $R_{det}$, and $F1_{det}$ represent the precision, recall, and F1 score of error detection. $P_{cor}$ represents the precision of correction on error tokens. The character error rate reduction (CERR) is also shown.}
\label{tab:deep_ana}
\vspace{-3mm}
\begin{center}
\scalebox{0.98}{
\begin{tabular}{l|ccccc|ccccc}
\toprule
\multirow{2}{*}{Model}   & \multicolumn{5}{c|}{AISHELL-1}  & \multicolumn{5}{c}{Aidatatang}
\tabularnewline\cmidrule{2-6}\cmidrule{7-11}
\multicolumn{1}{c|}{} &   $P_{det}$  &  $R_{det}$ & $F1_{det}$ & $P_{cor}$ & CERR  &   $P_{det}$  &  $R_{det}$ & $F1_{det}$ & $P_{cor}$ & CERR   \\\midrule
\multicolumn{10}{l}{\textit{Implicit error detection baselines}}\\\midrule
AR Correct  & \textbf{84.56} & 33.00 & 47.47 & 64.73 & 15.73 & 73.55 & 14.32 & 23.97 & 48.05 & 1.79 \\
AR N-Best  & 76.03 & 45.13 & 56.64 & \textbf{72.29} & 18.43 & 57.98 & \textbf{32.18} & \textbf{41.39} & 56.55 & -5.15 \\\midrule
\multicolumn{10}{l}{\textit{Explicit error detection baselines}}\\\midrule
FastCorrect  & 83.72 & 34.54 & 48.90 & 59.84 & 13.87 & 69.78 & 9.78 & 17.16 & 42.68 & 0.0 \\
FastCorrect 2 & 80.58 & 32.54 & 46.36 & 70.13 & 14.91 & 60.50 & 23.40 & 33.75 & 53.48 & -2.68\\\midrule
SoftCorrect  & 84.06 & \textbf{49.71} & \textbf{62.47} & 71.30 & \textbf{26.09} & \textbf{80.52} & 25.29 & 38.49 & \textbf{61.25} & \textbf{9.40}\\
\bottomrule
\end{tabular}}
\end{center}
\vspace{-2mm}
\end{table*}

\subsection{Method Analyses}
\label{PR_ana}

We compare the error detection and correction ability of SoftCorrect with previous autoregressive and non-autoregressive baselines. We measure the error detection ability using the precision ($P_{det}$) and recall ($R_{det}$) of that an error token is detected as error, and measure the error correction ability using the precision ($P_{cor}$) of that an error token is corrected to its corresponding ground-truth token. For autoregressive models that use implicit error detection, we assume a model detect a source token as error token if the model edits that token to another token.

As shown in Table~\ref{tab:deep_ana}, we can observe that: 1) on both datasets, SoftCorrect achieves better $P_{det}$, $R_{det}$, and $P_{cor}$ than non-autoregressive baselines with explicit error detection, which shows SoftCorrect has a stronger ability on error detection and correction; 2) Compared with autoregressive baselines using implicit error detection, SoftCorrect performs better on balancing the precision and recall of error detection, which verifies the necessity of soft error detection. 3) The errors in Aidatatang dataset is hard to detect, which cannot be handled with implicit error detection or duration-based explicit error detection. 
On this dataset, previous method may mistake a correct token which introduces new error, or miss an incorrect token. In contrast, the precision or recall of the detection in SoftCorrect is higher, demonstrating the advantage of soft error detection. Moreover, with high-accurate error detection, constrained CTC loss makes the error correction more focused and thus easier, resulting in the higher $P_{cor}$ of SoftCorrect.

\section{Conclusion}
\label{sec:conclusion}
In this paper, we design a soft error detection mechanism for ASR error correction, which consists of an encoder for error detection and a decoder for focused error correction. Considering error detection is important for ASR error correction and previous works using either explicit or implicit error detection suffer from some limitations, we propose SoftCorrect with a soft error detection mechanism. Specifically, we design an anti-copy language model loss to enable the encoder to select a better candidate from multiple input candidates and detect errors in the selected candidate, and design a constrained CTC loss to help decoder focus on correcting detected error tokens while keeping undetected tokens unchanged. Experimental results show that SoftCorrect achieves much larger CER reduction compared with previous explicit and implicit error detection methods in ASR error correction, while still enjoying fast inference speed.

\bibliography{aaai23}

\appendix

\section{Appendix}

\subsection{Pseudo Data Generation for Pretraining Correction Models}
\label{appen:data_gen}

Since ASR systems usually achieve good WER (e.g., $<$10\%), most tokens in the generated candidates are correct, which limits the number of effective training samples for error correction models. Therefore, a common practice \cite{leng2021fastcorrect, leng2021fastcorrect2,linchen2021improve,du2022crossmodal} is to construct pseudo error sentences from the raw text data to pre-train the correction model. Instead of manually constructing the pseudo error sentences by randomly deleting, inserting, and substituting tokens in text data~\cite{omelianchuk-etal-2020-gector,leng2021fastcorrect,wang-etal-2018-hybrid}, we leverage a BERT~\cite{devlin2019bert} model to automatically generate error tokens. Specifically, 1) during the training of this BERT generator, we introduce homophone substitutions to simulate ASR errors following~\citet{zhang-etal-2021-correcting}; 2) when generating error tokens, we convert some tokens to mask symbols and use the output of BERT generator as pseudo error tokens. The sentences with error tokens generated by BERT and their corresponding ground-truth sentences make up the pseudo dataset for ASR correction model. The BERT generator is only used for pretraining data generation and not jointly trained with our correction model.

The details of training of BERT Model are as follows. We train a BERT model with a masked language modeling loss~\cite{devlin2019bert}, by adding noise to some tokens in the input sentence (the ratio of adding noise is 20\%) and reconstructing the noised tokens. Given that a large proportion of ASR errors are homophones, in order to simulate this error pattern~\cite{zhang-etal-2021-correcting}, the strategy of adding noise is designed as: 1) 10\% tokens are unchanged; 2) 40\% tokens are replaced as a special $<mask>$ token; 3) 40\% tokens are replaced as its homophone; 4) the remaining 10\% tokens are replaced by random tokens. During the BERT training, we will first sample 20\% token which are going to be noised. Then we will sample strategy of adding noise based on the above 10\%-40\%-40\%-10\% distribution and add noise correspondingly. Our BERT model is a 12-layer Transformer model, where the attention head, hidden size, and filter size of each transformer layer are set as 8, 512, and 2048 respectively.

The details of generation are as follows. We use the trained BERT model to generate pseudo ASR error data with deletion, insertion, and substitution errors, which correspond to three kinds of errors caused by ASR system. Given a raw text sentence: 1) for deletion error, we random delete a token from sentence; 2) for insertion error, we random insert a mask symbol into sentence and feed it into BERT model to generate the inserted token; 3) for substitution error, we random replace a token in sentence with a mask symbol and feed it into BERT model to generate the substitution token. As a result, we can obtain three different kinds of ASR errors, which can be used to pre-train correction model.

Since there is no acoustic realization during pretraining on pseudo data, the acoustic scores are not used in pretraining stage.

\subsection{Encoder Variants in Ablation Studies}
\label{appen:enc_ablation}
In this section, we introduce the details of our encoder variants in ablation studies. In general, the encoder outputs a probability for each token, which will be further used for candidate selection. The training detail of setting ID 2, 3, 4, 5 in Table \ref{tab:ablation} are as follows.

\paragraph{Standard CE Loss}
This setting is similar with our SoftCorrect but the $GT$ token and its corresponding loss is removed, degenerating into a standard cross-entropy loss. It is equivalent to the first term of Equation 1 without the $GT$ token.

\paragraph{BERT-style}
Given aligned multiple candidates as input, we use masked language modeling loss in BERT to train the encoder. Specifically, we randomly replace all the tokens in some positions to mask symbols and let the model to predict their corresponding target token on that position. During inference, we infer $N$ ($N$ is the number of positions) times, where we mask one position in each time and obtain its probability.

\paragraph{GPT-style}
Given aligned multiple candidates as input, we use language modeling loss in GPT to train the encoder, i.e., the encoder is trained to predict next token based on previous tokens. Note that this setting only makes use of uni-directional information instead of bi-directional information as used in other encoder variants.

\paragraph{Binary}
We apply a binary classification loss on each token of each candidate on top of the encoder. The encoder is trained to classify whether the input token (i.e., each token in each candidate) is the target token or not (i.e., correct or incorrect).

\subsection{Further Discussion of SoftCorrect}
\label{appen:further_dis}
\paragraph{Relationship with Deliberation Models} Deliberation models \cite{hu2020deliberation,hu2021transformer,9747144} are another important kinds of correction models for ASR to correct errors by leveraging audio and first-path ASR output to generate a second-path result with less errors. We have tried deliberation model \cite{hu2020deliberation} but only to find that the deliberation model fails to correct the errors of ASR systems. The reason for the inferior accuracy of deliberation model is the insufficiency of training data on public ASR datasets. Since deliberation models need to implicitly align audio with text to detect and correct ASR errors, it requires more training data. To the best of our knowledge, most of works on deliberation models are trained and verified only on private product data of big companies \cite{hu2020deliberation,hu2021transformer,9747144}. The data size of this private product data is about 100K hours, which is hundreds or thousands of times bigger than that of public ASR datasets (i.e., 100-1000 hours).

The deliberation models and text-based correction models (e.g. proposed model and its baselines) are suitable for different scenarios and are both meaningful. Deliberation models are better when both ASR data and text data are quite abundant (e.g., private product datasets of big company on popular languages). However, it is rare that we have an extremely large ASR dataset. In most of conditions (such as ASR of low-resource languages and public ASR datasets), the audio data is limited but the unpaired text data is abundant since the unpaired text data is generally much easier to collect compared with the ASR data. And the text-based correction models are suitable for this condition.

\paragraph{The Robustness of Error Corrector}

It is important for the corrector to be robust when the error detector makes a mistake. Considering that failing to detect an error token will not introduce new errors but mistakenly editing a correct token will introduce new errors, we use the following designs to make corrector more robust if the detector mistakenly detects a correct token as error token.

When training corrector, we randomly select 5\% correct tokens and regard them as pseudo error tokens (to simulate the detection mistake from detector). Those pseudo error tokens will be repeated and optimized back to themselves (since they are actually correct and do not need correction). As a result, the corrector will be robust to the mistake from detector.

The above mechanism eases the problem when the detector mistakenly detects a correct token as error. Another type of error is that the detector might fail to detect an error token. However, as Section \ref{PR_ana} shows, the recall of error detection is lower than 50\%, which shows that it is inevitable for the detector to fail to detect a proportion of error tokens. We just let this error propagate to error correction stage by not changing tokens detected as correct, since 1) it is hard to detect all errors, 2) failing to edit an error token will not lead to bad consequence (i.e., introducing new errors).

\paragraph{The Completeness of SoftCorrect}
For the hardest part of deletion error (i.e., the left and right neighbor token of the deletion error token are both correct, such as the ASR output is "AC" while the ground-truth token is "ABC"), SoftCorrect has a limitation on the detection of it. However, if 1) one of the neighbor tokens also has error or 2) the deleted token is predicted in any of the multiple candidates, SoftCorrect is possible to detect it out. For a correction model, the deletion error is hard to detect and harder to correct since there is almost no clue about the missed token. Giving up this part of errors might not lead to performance drop. The completeness of SoftCorrect can be improved by adding an additional binary error detector for this kind of error.

\paragraph{Future Work} For future work, we will explore SoftCorrect in other similar tasks like grammar error correction and text edition, and support more scenarios like streaming input.

\begin{table}[t]
\caption{Results on WenetSpeech Dataset.}
\label{tab:wenet}
\begin{small}
\begin{center}
\scalebox{0.88}{
\begin{tabular}{l|c|c|c|c|c}
\toprule
 \multicolumn{2}{c|}{\multirow{3}{*}{Model}} & \multicolumn{2}{c|}{Test Meeting}  & \multicolumn{2}{c}{Test Net} \\
\cmidrule{3-6} 
\multicolumn{2}{c|}{} & CER $\downarrow$  & CERR $\uparrow$ &  CER $\downarrow$  & CERR $\uparrow$\\
\midrule
 \multicolumn{2}{l|}{No Correction} &  15.61 & -  & 8.74 & - \\
\midrule
 \multicolumn{6}{l}{\textit{Implicit error detection baselines}} \\
 \midrule
 \multicolumn{2}{l|}{AR Correct} &  15.40 & 1.35  & 8.65 & 1.03\\
\midrule
 \multicolumn{2}{l|}{AR N-Best} &  15.45 & 1.02  & 8.89 & -1.72 \\
\midrule
 \multicolumn{6}{l}{\textit{Explicit error detection baselines}} \\
 \midrule
 \multicolumn{2}{l|}{FastCorrect} &  15.46 & 0.96  & 8.68 & 0.69\\
\midrule
 \multicolumn{2}{l|}{FastCorrect 2} &  15.49 & 0.77  & 9.11 & -4.23 \\
\midrule
 \multicolumn{6}{l}{\textit{Other baselines}} \\
 \midrule
 \multicolumn{2}{l|}{Rescore} & 14.82 & 5.06  & 8.58 & 1.83 \\
\midrule
 \multicolumn{2}{l|}{Rescore + FC} &  14.60  & 6.47 & 8.48 & 2.97 \\
\midrule
 \multicolumn{2}{l|}{FC + Rescore} &  14.65  & 6.15 & 8.51 & 2.63 \\
\midrule
 \multicolumn{2}{l|}{SoftCorrect} &  \textbf{14.09}  & \textbf{9.74} & \textbf{8.26} & \textbf{5.49} \\

\bottomrule
\end{tabular}}
\end{center}
\end{small}
\end{table}

\begin{table*}[t]
\caption{The CERR of SoftCorrect with respect to the CER interval of ASR output. ``No Corr CER'' stands for the CER of an interval without correction. ``Corr CER'' stands for the CER of an interval after the correction of SoftCorrect.}
\label{tab:p_wrt_wet}
\begin{small}
\begin{center}
\scalebox{1}{
\begin{tabular}{l|c|c|c|c|c|c|c}
\toprule
 \multicolumn{2}{c|}{\multirow{3}{*}{CER Interval}} & \multicolumn{3}{c|}{AISHELL-1}  & \multicolumn{3}{c}{Aidatatang} \\
\cmidrule{3-8} 
\multicolumn{2}{c|}{} & No Corr CER  & Corr CER  & CERR & No Corr CER  & Corr CER  & CERR\\
\midrule
 \multicolumn{2}{c|}{(0, 5]} &  4.53  & 2.68 & 40.88 & 4.12  & 3.32 & 19.33 \\
\midrule
 \multicolumn{2}{c|}{(5, 10]} &  6.96  & 4.34 & 37.63 & 7.24  & 5.88 & 18.83 \\
\midrule
 \multicolumn{2}{c|}{(10, 15]} &  11.86  & 7.83 & 33.98 & 11.97  &10.00 & 16.45 \\
\midrule
 \multicolumn{2}{c|}{(15, 20]} &  16.83  & 11.86 & 29.53 & 16.97  & 14.36 & 15.38 \\
\midrule
 \multicolumn{2}{c|}{(20, $+\infty$)} &  27.30  & 21.32 & 21.89 & 33.14  & 29.87 & 9.88 \\

\bottomrule
\end{tabular}}
\end{center}
\end{small}
\end{table*}

\subsection{Training Configurations}
\label{appen:training_details}
We use Transformer~\cite{vaswani2017attention} as the backbone of our encoder and decoder, where the encoder (detection module) and the decoder (correction module) have respectively 12 and 6 layers, where hidden size is 512. The hyper-parameter $\lambda$ in anti-copy language model loss is set to 1.0. We trained SoftCorrect and the baseline systems on 16 NVIDIA V100 GPUs, with a batch size of 6000 tokens. These systems are first pre-trained with 400M pseudo text data and then fine-tuned on paired training data, i.e., ASR transcription and ground-truth target. We measure the inference speed, i.e., latency, of each system on an NVIDIA V100 GPU. For other hyper-parameters, we follow the setting in Fairseq~\cite{ott2019fairseq}\footnote{github.com/pytorch/fairseq/tree/main/examples/translation}.

\subsection{Results on WenetSpeech Dataset}
\label{appen:wenetspeech}
To verify the effectiveness of SoftCorrect on larger dataset, we conduct experiment on WenetSpeech dataset \cite{wenetspeech}, which is, to the best of our knowledge, the largest (10000-hour) public available ASR dataset. The ASR model used in our experiment is following ESPnet \cite{watanabe2018espnet} \footnote{github.com/espnet/espnet/tree/master/egs2/wenetspeech}.
The results on two test set of WenetSpeech (Test Metting and Test Net) are shown in Table \ref{tab:wenet}.

The results on WenetSpeech are consistent with that on AISHELL-1 and Aidatatang (as shown in the Section \ref{subsec:acc_latency}), where SoftCorrect surpasses all baseline systems by a margin and achieves 9.74\% and 5.49\% CERR on Test Meeting and Test Net set of WenetSpeech, respectively, which verifies that SoftCorrect is effective when the ASR dataset grows to 10000 hours.

\subsection{Significance Tests}
\label{appen:sig_test}
For significance tests, we conduct MAPSSWE tests\footnote{github.com/talhanai/wer-sigtest}. The results show that the difference of SoftCorrect with all baseline methods are  statistical significant (p$<$0.005) in Table \ref{tab:main_result}. The only exception is SoftCorrect and "FC+R" baseline achieve comparable results on AISHELL-1 dataset (p=0.968) but being 7 times faster.

\subsection{Performance with respect to the CER of ASR System}
\label{appen:p_wrt_wet}
Since not every ASR model has a low CER in the first pass, it is important to understand the performance of SoftCorrect when the ASR model has a relative inferior accuracy. To understand this, we split the test set to 5 intervals based on the CER of ASR output and calculate the corresponding CERR of each interval, which is shown in Table \ref{tab:p_wrt_wet}.

From the Table \ref{tab:p_wrt_wet}, we observe that with the increase of CER of ASR output, the absolute drop of CER (i.e., from "No Corr CER" to "Corr CER") by our correction is increasing. We also notice the CERR of SoftCorrect increases as the CER of ASR output becomes lower. The ability of SoftCorrect on correcting large \textit{number} of errors when the CER of ASR output is high and correcting large \textit{proportion} of errors when the CER of ASR output is low verifies the effectiveness of SoftCorrect.

\end{document}